\documentclass{article}
\usepackage{PRIMEarxiv}
\usepackage[utf8]{inputenc} 
\usepackage[T1]{fontenc}    
\usepackage{hyperref}       
\usepackage{url}            
\usepackage{booktabs}       
\usepackage{amsfonts}       
\usepackage{nicefrac}       
\usepackage{microtype}      
\usepackage{lipsum}
\usepackage{fancyhdr}       
\usepackage{graphicx}       
\usepackage{microtype}
\usepackage{graphicx}
\graphicspath{ {./images/} }
\usepackage{latexsym}
\usepackage{amsmath}
\usepackage{times}
\usepackage{epsfig}
\usepackage{amssymb}
\usepackage[export]{adjustbox}
\usepackage{array}
\usepackage{enumitem}
\usepackage[ruled,vlined]{algorithm2e}
\DeclareMathAlphabet{\pazocal}{OMS}{zplm}{m}{n}

\usepackage{tabularx}               
\newcolumntype{C}{>{\centering\arraybackslash}X}
\usepackage{multirow}               
\usepackage{diagbox}                
\usepackage{hhline}                 
\usepackage{color}                  
\usepackage{amssymb}                
\usepackage{mathtools}              
\usepackage{enumitem}               

\usepackage{makecell}
\usepackage{xcolor}

\definecolor{effectspancolor}{RGB}{0, 51, 125}
\definecolor{drugspancolor}{RGB}{44, 7, 110}
\usepackage{xparse}

\title{Curriculum Learning for Language Modeling}

\author{Daniel Campos \thanks{Work done pursing Masters Degree at University of Washington}\\ 
University of Illinois Urbana-Champaign \\ 
\texttt{dcampos3@illinois.edu} \\
  }

\begin{document}
\maketitle

\begin{abstract}
Language Models like ELMo and BERT have provided robust representations of natural language, which serve as the language understanding component for a diverse range of downstream tasks.Curriculum learning is a method that employs a structured training regime instead, which has been leveraged in computer vision and machine translation to improve model training speed and model performance. While language models have proven transformational for the natural language processing community, these models have proven expensive, energy-intensive, and challenging to train. In this work, we explore the effect of curriculum learning on language model pretraining using various linguistically motivated curricula and evaluate transfer performance on the GLUE Benchmark. Despite a broad variety of training methodologies and experiments we do not find compelling evidence that curriculum learning methods improve language model training.
\end{abstract}
\section{Introduction}
Seeking to represent natural language, researchers have found language models (LM) with Sesame Street-inspired names \cite{Peters2018DeepCW} \cite{Devlin2019BERTPO} \cite{Sun2019ERNIEER} to be incredibly effective methods of producing language representations (LR). These LM's have leverage transfer learning by training on a large text corpus to learn a good representation of language which can then be used in a down steam task like Question Answering or Entity Resolution. While these LMs have shown to be excellent methods to enable language understanding, the ability to train these models is becoming increasingly computationally expensive \cite{Strubell2019EnergyAP}. Since model performance is closely tied to the size of training data, model size, and compute used to train \cite{Kaplan2020ScalingLF} the bulk of existing research has focused on scaling these aspects without much focus on increasing efficiency of training. Seeking to explore what methods could be used to make LM training more efficient we study the effect of curriculum learning by training ELMo with a wide variety of curricula. \\
Curriculum learning (CL) is a training methodology which applies structure to a models training data. CL has been studied broadly in natural language processing and has been very successful in domains like Neural Machine Translation (NMT) where CL based models are able to train faster and produce better results \cite{Platanios2019CompetencebasedCL} \cite{Zhang2019CurriculumLF} \cite{Penha2020CurriculumLS} than unstructured, stochastic sampling. Focusing on LMs, Xu et al. \cite{xu-etal-2020-curriculum} showed that CL can be used in LM finetuning as a way to improve task performance. Despite an abundance of work exploring CL and LMs to the best of our knowledge we are the first to examine the effect of curriculum learning in LM pre-training and transfer performance. \\
To evaluate the effect of CL on LMs we train ELMo with a variety of curricula on the wikitext-2 and wikitext-103 \cite{Merity2016PointerSM} without modification of training time or model hyperparameters. We evaluate model performance on the pre-training task and on the GLUE Benchmark \cite{Wang2018GLUEAM} building on the work of Competence Based Curriculum Learning \cite{platanios-etal-2019-competence} by modifying training sampler within the LM to produce a dataset with gradually increasing difficulty \footnote{Code and results available at https://github.com/spacemanidol/CurriculumLearningForLanguageModels }. The contributions of our work are:
\begin{itemize}
  \item Exploration of the effects of curriculum learning for language modeling finding no clear improvement to models that use curriculum methods for training.
  \item Experiments suggesting random curriculum in which the structure of the training regime is random can be just as effective as linguistically motivated methods.
\end{itemize} 
\section{Related Work}
\subsection{Curriculum Learning}
CL subset of training regimes which introduce structure to improve training efficiency, model performance, or model model robustness by optimizing what kind of information a model has access at each training step. Experiments with RNNs \cite{Elman1993LearningAD} suggested that learning of complex grammatical structure improves when the initial examples the models learn with are more easier. Experiments in modifying language modeling training data find a lower loss can be achieved by training on incrementally more difficult data \cite{Bengio2009CurriculumL}. Recently, competence based curriculum \cite{Platanios2019CompetencebasedCL} has been used to improve machine translation progressively modifying the training corpus until it matches the original distribution. It has been used to reduce training time by up to 70\% and improve BLEU performance by 2.2 points on the WMT dataset. For further readings about curriculum learning, applications and current bottlenecks we recommend Soviany et al.'s survey \cite{Soviany2021CurriculumLA}
\subsection{Language Modeling}
Language modeling is a way to assign a probability distribution over some textual representation. In other words, if the task is to model $n$-grams, the probability of a current input is the probability of a token $w_i$ given the previous $i$ tokens. Language Models like ELMo \cite{Peters2018DeepCW} and BERT \cite{Devlin2019BERTPO} leverage large text corpora to learn language representations that can be used for downstream tasks like text classification or question answering. While LMs lead to large improvement in performance for downstream tasks they are both expensive and complex to train. A single training run of a model like GPT-2 can cost upward of \$40,000, the architecture search and hyperparameter tuning can be upwards of \$3,000,000, and the C$0_2$ released by training one of these models can be similar to the C$0_2$ released in the entire life-cycle of a car \cite{strubell-etal-2019-energy}. 
\section{Method}
Language modeling is a way to assign a probability distribution over some textual representation. This probability distribution is commonly modeled as the probability of a current token $w_i$ given the previous $i$ tokens as formally represented in equation \ref{equation:langmodel}. Using language modeling as a pre-training method, LMs learn representations which can be used in downstream tasks. Since language has structure, we believe that structuring the pre-training methodology can lead to improved model performance. To introduce structure into LM training we leverage Platanios et al.'s competence based curriculum (CBC)\cite{Platanios2019CompetencebasedCL} as shown in Algorithm 1. CBC uses a notion of model competence and sample difficulty to control what a model learns. First, the corpus, $X$, a collection of samples $S$, where each sample $s_i$ is a sequence of words $s_i= w_o^i,w_1^i,...,w_n^i$ is sorted by difficulty using a which Using a heuristic like sentence length or unigram rarity is assigned a difficulty $\epsilon_{s_i}=[0,1]$. Given a processed corpus, a model is assigned a initial competence $\lambda_0$ and a competence increment $lambda_{increment}$. A model's competence score is a representation of how far along in a training regime the model is. At each training step, a model samples from data that is lower than its current competence, updates its weights, and increases its competence. The model is only able to train on samples that have a difficulty where $\epsilon_{s_i} <= \lambda_t$. \\
\begin{algorithm}[h]
\small
\label{algo:competence}
\SetAlgoLined
\KwResult{Model Trained with Competence Based Curriculum}
Input: X, $\lambda_0$, $\lambda_{increment}$ \;
Compute difficulty, $\epsilon_{s_i}$ for $s_i \in X$\;
Compute Cumulative density of $\epsilon_{s_i}$\;
$\lambda_t = \lambda_0$\;
\For{training step t = 1,...,n}{
Sample batch $b$ from X such that $\epsilon_{s_i} < \lambda_t$\;
Train on batch $b$\;
$\lambda_{t+1} = \lambda_t + \lambda_{increment}$\;
}
\caption{CBC Training Regime}
\end{algorithm}
Using CBC we explore 8 proxies for sample difficulty: no curriculum, random, sample length, unigram sample probability, bigram sample probability, trigram sample probability, part of speech diversity (POS), and sample dependency parse complexity (DEP). For each methodology, for each $s_i$ in $X$, we compute a difficulty value for each sample $\epsilon_{s_i}$ and then sort the dataset by this difficulty score. Using the sorted dataset we compute the cumulative density function (CDF) giving each sample of the difficulty score $\epsilon_{s_i} \in [0,1]$. No curriculum sets $\lambda_0 = 1$ which means training samples stochastically and serves as a baseline. A random curriculum is generated by assigning values for $\epsilon_{s_i}$ at random and establishes the effect of any arbitrary structure. The remaining six heuristics are based on common NLP difficulty metrics and linguistically motivated heuristics. 
\subsubsection{Sample Length}
Sample Length builds on the idea that is a lot harder to model longer sentences, as longer sentences require better tracking of dependencies. It is calculated by creating a CDF on $\text{sentence-length-}\epsilon_{s_i} = length(s_i)$.\\ 
\subsubsection{Sentence Entropy: $N$-gram difficulty}
Sentence Entropy builds uses the notion that can be difficult to model is words with a variety of frequency in the corpora. Models, if assumed to behave like humans, would find it difficult to understand the meaning of a word if they do not see it in a corpus nor have a diversity of usages to infer meaning. Since the statistical strength of training samples with rare words is low and the early model learned word embeddings are likely to have high variance it is likely that exposing a model early to rare words can result in badly estimated representations. To quantify this difficulty we propose producing a sentence entropy for each sentence with respect to its unigram, bigram, and trigram probabilities. These are calculated using standard sample entropy calculations as shown below
Sample entropy for each $N$-gram can be thought of the probability of the sample occurring given an approximate naive language modeling assuming words are sampled independently. Samples are scored by calculating the product of $n$-gram log likelihood given the sample. Note, we are not calculating the conditional probability of each word given the preceding N words but the probability of the N-gram given the text corpus. Calculation of $\epsilon_{s_i}$ is shown in equation \ref{equation:gramprob} where $uc$, $bc$, and $tc$ are the counts of unique unigrams, bigrams, and trigrams in the corpus, $C$ is the corpus, $c(y)$ is the count of $y$ in a sample, $x \in C$ is a sample in the corpus and $w_i \in x$ is a word in a line, and $l(x)$ is the length of $x$ in $n$-grams.
\begin{equation}
\begin{split}
    p(w_n) &= \frac{\sum_{x \in C} c(w_n)}{\text{us}} \\
    p(w_{n}, w_{m})  & = \frac{\sum_{x \in C} c(w_n,w_m)}{\text{bs}} \\
    p(w_{n}, w_{m}, w_{j})  & = \frac{\sum_{x \in C} c(w_n, w_m, w_j) }{\text{ts}} \\
    \text{unigram-} \epsilon({s_i}) &= \prod_{n=0}^{l(s_i)} \log(p(w_n)) \\
       \text{bigram-} \epsilon({s_i}) &= \prod_{n=0}^{l(s_i)-1} \log(p(w_{n-1}, w_n)) \\
       \text{trigram-} \epsilon_{s_i} &= \prod_{n=0}^{l(s_i)-2} \log(p(w_{n}, w_{n+1}, w_{n+2})) 
\end{split}
\label{equation:gramprob}
\end{equation}
\subsubsection{Dependency Tree}
Sentences are often modeled as dependency trees to model the interaction between words and groups of words in a text sample. While not infallible, sentences that have a deeper tree usually more complex and as a result more difficult. We leverage the language processing framework SPACY.IO's to generate parse trees for each sample and measure the depth of each tree. This information is then used to calculate difficult as $\text{dep-}\epsilon_{s_i} = \text{depth}(s_i)$. Since there are fewer unique values for tree depth this method can be inferred to have a high commonality with random difficulty.\\
\subsection{Part of Speech Diversity}
Another core part of language complexity can be derived by the diversity of parts-of-speech in a sentence. We believe that more difficult sentences feature a higher diversity of parts-of-speech (POS) and use SPACY.IO's part of speech tagger to produce a set for in each sample and calculate difficulty with $\text{pos-}\epsilon_{s_i} = \text{len}(\text{set}(\text{pos}({s_i})))$
\begin{equation}
\begin{split}
    P(w_{1},\ldots ,w_{m}) &= \prod _{i=1}^{m}P(w_{i}\mid w_{1},\ldots ,w_{i-1})\\
    & \approx \prod _{i=1}^{m}P(w_{i}\mid w_{i-(n-1)},\ldots ,w_{i-1})
\end{split}
\label{equation:langmodel}
\end{equation}
\section{Experiments}
To evaluate the effect of curriculum learning on language modeling we train ELMo models varying the training corpus and using our aforementioned difficulty proxies to generate various language models. Training leverages well-established language modeling benchmarks of wikitext-2, and wikitext-103 \cite{Merity2016PointerSM} with details can be found in table \ref{table:corpussize} . These datasets collect verified good and featured articles from English Wikipedia and feature 2 million and 103 million tokens and were selected for the variations in size and speed of training. After training, each language model performance is then evaluated based on performance on the training corpus (measured in perplexity) and transfer ability on  The General Language Understanding Evaluation Benchmark (GLUE) \cite{Wang2018GLUEAM}. GLUE is a set of resources focused on the evaluation of natural language understanding systems. This benchmark pools eleven sentence-level language understanding tasks tasks.
\begin{table}[h!]
\resizebox{\columnwidth}{!}{\begin{tabular}{|l|l|l|l|l|} \hline
\textbf{Corpus Name} & \textbf{vocabulary Size} & \textbf{Tokens} & \textbf{lines} & \textbf{sentences} \\ \hline
wikitext=2 & 33278 & 2507007 & 44836 & 131262 \\ \hline
wikitext-103 & 267735 & 103690236 & 1809468  & 5343947 \\ \hline
1B Word Benchmark & 793471 & 829250940 & N/A & N/A \\ \hline
\end{tabular}}
\caption{Training Corpus details}
\label{table:corpussize}
\end{table} 
\subsection{Pre-Training Details}
Using the 16 curricula (8 for each corpus) we train an ELMo model using the original code \footnote{https://github.com/allenai/bilm-tf} with a modified batch sampler created for competence based sampling. For baselines, we train Elmo models without our modified CBC sampling using wikitext-2, wikitext-103. Following the original work, we train each curriculum-model variant for 10 epochs on the pre-training corpus, use 2 stacked 4096 dimensional BiLSTMs, use dropout of 0.1, batch size of 128, and a context window of 20 tokens. Training was performed using 3 Nvidia 2080 TI GPUs requiring about 30 hours for the wikitext-103 and about an hour for wikitext-2. For CBC training hyperparameters, we performed a grid search on $\lambda_{increment}$ and $\lambda_0$ values finding the lowest training perplexity at $\lambda_0 = 1e-1$  $\lambda_0 = 1e-1$ for wikitext-2 and $\lambda_0 = 1e-3$ $\lambda_{increment} = 1e-5$ for wikitext-103. \\
In the original implementation, the training loader loads a file, shuffle all the lines, and samples batches by iterating through the shuffled corpus. Our method load the full corpus, then select a batch at random from the examples that meet our model's current competence. This changes data sampling to unconstrained random sampling without replacement to sampling with replacement. Since our competence based sampling has a variety of sample lengths we use the padding token $<PAD>$ as is common in NMT. All samples are padded to length 20 and the loss on these padding tokens is set to zero. Since padding tokens for wikitext-103 we introduce 12,204,311 tokens equating to approximately 12 percent more FLOPs.
\subsection{Transfer Learning}
After models have pretrained we evaluate on GLUE performance using the JIANT toolkit \cite{Pruksachatkun2020jiantAS}. JIANT is an open-source tool for conducting multi-task and transfer learning experiments in English to implement the GLUE benchmark. JIANT builds on the notion of a configuration which provides all settings needed to run and reproduce an experiment in a simple text file. JIANT provides consistent data processing, classifier implementation, and evaluation to ensure that users of the framework can focus on the outputs and not worry about implementing benchmarking tasks like GLUE. JIANT uses the pretrained model weight along with a multi-layer perceptron with 512 hidden dimensions to train on each GLUE tasks. Each JIANT experiment fixes training identically across tasks and inputs using a batch size of 8, random seed of 42 initial learning rate of 1e-1, dropout of 0.2. Training of each model continues until the model learning rate dips below 1e-6 or the model performance has not improved in 10 epochs. Unless another metric is explicitly mentioned, the GLUE sub-task metric is accuracy. 
\subsection{Experimental Results}
Focusing on model pretraining performance, despite attempts in variation of $\lambda_0$ and $\lambda_{increment}$, no implementation of CBC is able to approximate the baselines in term of perplexity on the held out portion of the wikitext-* dataset. Complete graphs can be found in the appendix but all curricula perplexities including the baseline are order of magnitudes higher than stochastic sampling. On wikitext-2, the best performance is achieved by the curricula baseline ($\lambda_0=1$) with a perplexity of 770, followed by random with a perplexity of 2105 well above the baseline of 151. We believe this is caused by the change in dataset distribution caused by our curriculum learning implementation. Similar effects are seen on wikitext-103 where unlike stochastic sampling, which achieve a perplexity of 36, curriculum methods are unable to achieve a perplexity under one thousand. Surprisingly, as data size scales we see a larger volatility in perplexity during training with respect to validation perplexity scores which we attribute to constantly shifting sample distribution caused by curriculum methods. \\
As we move our focus to GLUE results on wikitext-2 based models in table \ref{tab:wiki2-glue}, we find that curriculum methods generally outperform stochastic sampling by 10\%. We do not find strong evidence that the structure of the curriculum matters as non curriculum ($\lambda_0=1$) performs better than 4 other curricula and the baseline. Perhaps most surprising, random outperforms the baseline when measure by overall glue score despite their being no formal structure in the training regime. Observing variability at an individual task level we find only CoLA, STS-B and SST have broad variability in performance. We believe this is because these tasks are smaller and more linguistically challenging.\\
Focusing on results on larger corpus in table \ref{tab:glue-wiki-103} we find trends found in wikitext-2 no longer hold as top performance is achieved by stochastic unmodified baseline. We also note that the orderign of system performance does not hold across datasets and as the pretraining dataset grows the variability between models decreases. Similar to the smaller corpus, we find the highest sensitivity in ColA and find variability in SST and STS-B has become more muted. Surprisingly, given their had the worse performance in pretraining perplexity, the trigram curricula generate the best transfer task.  \\
Overall, we find that CBC training provided worse performance on validation perplexity portion and improved performance on transfer tasks when he pretraining corpus is small. We believe this reinforces the importance of the size of the pretraining corpus since a large corpus allows the model to learn better language representations without any structured training. We also find a large disconnect in model pretraining perplexity and transfer task performance as performance on one is not predicative of the other.
\begin{table}[h]
\resizebox{\columnwidth}{!}{%
\begin{tabular}{|l|l|l|l|l|l|l|l|l|l|l|l|}
\hline
Method & Overall & Cola & SST & MRPC & STS-B & QQP & MNLI & QNLI & RTE & WNLI & DX \\\hline
dep & \textbf{0.63} & \textbf{0.19} & 0.73 & 0.85/0.78 & \textbf{0.71/0.71} & 0.74/0.78 & 0.60 & 0.75 & 0.58 & \textbf{0.56} & 0.11 \\\hline
unigram & \textbf{0.63} & 0.18 & 0.77 & \textbf{0.86/0.78} & 0.68/0.67 & \textbf{0.74/0.79} & 0.60 & 0.75 & 0.56 & 0.54 & 0.13 \\ \hline
trigram & \textbf{0.63} & 0.15 & 0.76 & 0.84/0.76 & 0.70/0.69 & 0.73/0.78 & \textbf{0.62} & \textbf{0.76} & 0.54 & \textbf{0.56} & 0.14 \\\hline
length & \textbf{0.63} & \textbf{0.19} & 0.75 & 0.84/0.77 & 0.66/0.65 & 0.73/0.78 & 0.60 & 0.75 & 0.57 & \textbf{0.56} & 0.13 \\ \hline
no curricula & 0.62 & 0.15 & 0.75 & 0.84/0.77 & \textbf{0.71/0.71} & 0.73/0.78 & 0.61 & 0.72 & 0.54 & \textbf{0.56} & 0.12 \\\hline
bigram & 0.62 & 0.18 & \textbf{0.77} & \textbf{0.86/0.78} & 0.68/0.67 & \textbf{0.74/0.79} & 0.60 & 0.75 & 0.56 & 0.44 & 0.13 \\\hline
random & 0.61 & 0.00 & 0.76 & 0.85/0.78 & 0.70/0.70 & 0.72/0.78 & 0.61 & 0.75 & 0.58 & \textbf{0.56} & 0.14 \\\hline
pos & 0.61 & 0.00 & 0.74 & 0.84/0.77 & 0.66/0.66 & 0.71/0.77 & 0.61 & 0.75 & \textbf{0.59} & \textbf{0.56} & \textbf{0.16} \\\hline
baseline & 0.59 & 0.00 & 0.70 & 0.85/0.78 & 0.66/0.66 & 0.70/0.75 & 0.59 & 0.72 & 0.54 & \textbf{0.56} & 0.13 \\\hline
length & 0.53 & 0.01 & 0.75 & 0.81/0.67 & \textbf{0.71/0.71} & 0.54/0.68 & 0.33 & 0.51 & \textbf{0.59} & 0.52 & 0.01 \\ \hline
\end{tabular}%
}
\caption{GLUE results for CBC models trained on wikitext-2. }
\label{tab:wiki2-glue}
\end{table}
\begin{table}[]
\centering
\resizebox{\columnwidth}{!}{%
\begin{tabular}{|l|l|l|l|l|l|l|l|l|l|l|l|}
\hline
Method & Overall & Cola & SST & MRPC & STS-B & QQP & MNLI & QNLI & RTE & WNLI & DX \\\hline
baseline & \textbf{0.67} & \textbf{0.28} & \textbf{0.86} & \textbf{0.87/0.80} & 0.77/0.77 & 0.72/0.76 & 0.64 & 0.76 & \textbf{0.61} & 0.54 & 0.14 \\\hline
trigram & 0.66 & 0.21 & 0.85 & \textbf{0.87/0.80} & \textbf{0.78/0.78} & \textbf{0.75/0.80} & \textbf{0.66} & \textbf{0.77} & 0.56 & 0.55 & 0.14 \\\hline
no curriculum & 0.66 & 0.21 & 0.83 & 0.87/0.8 & 0.77/0.77 & 0.75/0.79 & 0.64 & \textbf{0.77} & 0.58 & \textbf{0.56} & \textbf{0.15} \\\hline
bigram & 0.66 & 0.18 & 0.83 & 0.85/0.79 & 0.77/0.77 & 0.75/0.79 & 0.65 & \textbf{0.77} & 0.56 & \textbf{0.56} & 0.14 \\\hline
length & 0.66 & 0.21 & 0.82 & 0.85/0.72 & 0.77/0.77 & 0.73/0.79 & 0.63 & 0.75 & 0.58 & \textbf{0.56} & 0.14 \\\hline
unigram & 0.65 & 0.19 & 0.82 & 0.86/0.79 & 0.76/0.75 & 0.75/0.79 & 0.63 & 0.75 & 0.57 & \textbf{0.56} & 0.13 \\ \hline
random & 0.65 & 0.18 & 0.84 & 0.86/0.79 & 0.77/0.77 & \textbf{0.75/0.80} & 0.64 & \textbf{0.77} & 0.58 & 0.49 & 0.14 \\\hline
pos & 0.65 & 0.16 & 0.83 & 0.86/0.79 & 0.76/0.76 & 0.75/0.79 & 0.63 & 0.73 & 0.57 & \textbf{0.56} & 0.14 \\ \hline
dep & 0.64 & 0.23 & 0.85 & 0.86/0.78 & 0.78/0.78 & 0.75/0.79 & 0.64 & 0.76 & 0.54 & 0.42 & 0.14 \\ \hline
\end{tabular}%
}
\caption{GLUE results for CBC methods trained on wikitext-103.}
\label{tab:glue-wiki-103}
\end{table}
\subsection{Failure of Competence Based Curriculum}
In our experiments we were quite surprised at the failure to learn the training data found in our implementation of competence based curriculum as shown by the high perplexity on the wikitext-* datasets. Based on the changes in validation perplexity we believe the model is over-fitting on the altered training data. We believe the cause of this is our hyperparameter selection for $\lambda_0$ and $\lambda_{increment}$. We realize that since each method is effusively sampling from a different training distribution comparison of training perplexities are not comparable. Additionally, if we look at the difference in validation perplexity curves of various methods it is apparent that they are not learning at the same rate. Some methods like DEP, and POS do not see major fluctuations indicating the chosen curriculum parameters work well while many of the $n$-gram methods consistently fluctuate in a similar fashion indicating the chosen hyperparameters are sub optimal for them. Given the non trivial computational cost to explore $\lambda_0$ and $\lambda_{increment}$ for each method and the disconnect seen between pre-training perplexity and performance on GLUE we did not perform additional hyperparameter optimization.
\section{Conclusion and Future Work}
In our work we do not find strong evidence that the use of curriculum learning is able to improve language model pretraining. Our CBC based training regimes are unable to learn a good representation of the training corpus but their representations transfer well to downstream NLP tasks. We find that with a small pretraining corpus, CBC methods can outperform stochastic sampling but as corpus size scales the benefit is lost. Moreover we do not find any evidence that any type of heuristic for difficulty to be more apt for CBC. \\
Leveraging recent work on training BERT in an academic setting \cite{Izsak2021HowTT} we will explore how our top performing methods like trigram and random perform when their code becomes public. Building on promising results in model compression, \cite{Frankle2019TheLT} \cite{Wynter2020OptimalSE} we would like to explore how transfer learning effects compression. Inspired by the structure transformer-based models, we would explore jointly how progressively scaling the number of transformer encoders while increasing data difficulty, or context windows as this has proven useful for GANs \cite{Karras2017ProgressiveGO}.   
\bibliographystyle{unsrt}  
\bibliography{references}  

\begin{thebibliography}{10}

\bibitem{Peters2018DeepCW}
Matthew~E. Peters, Mark Neumann, Mohit Iyyer, Matt Gardner, Christopher Clark,
  Kenton Lee, and Luke Zettlemoyer.
\newblock Deep contextualized word representations.
\newblock {\em ArXiv}, abs/1802.05365, 2018.

\bibitem{Devlin2019BERTPO}
Jacob Devlin, Ming-Wei Chang, Kenton Lee, and Kristina Toutanova.
\newblock Bert: Pre-training of deep bidirectional transformers for language
  understanding.
\newblock In {\em NAACL-HLT}, 2019.

\bibitem{Sun2019ERNIEER}
Yu~Sun, Shuohuan Wang, Yukun Li, Shikun Feng, Xuyi Chen, Han Zhang, Xin Tian,
  Danxiang Zhu, Hao Tian, and Hua Wu.
\newblock Ernie: Enhanced representation through knowledge integration.
\newblock {\em ArXiv}, abs/1904.09223, 2019.

\bibitem{Strubell2019EnergyAP}
Emma Strubell, Ananya Ganesh, and Andrew McCallum.
\newblock Energy and policy considerations for deep learning in nlp.
\newblock In {\em ACL}, 2019.

\bibitem{Kaplan2020ScalingLF}
Jean Kaplan, Sam McCandlish, Tom Henighan, Tom~B. Brown, Benjamin Chess, Rewon
  Child, Scott Gray, Alec Radford, Jeffrey Wu, and Dario Amodei.
\newblock Scaling laws for neural language models.
\newblock {\em ArXiv}, abs/2001.08361, 2020.

\bibitem{Platanios2019CompetencebasedCL}
Emmanouil~Antonios Platanios, Otilia Stretcu, Graham Neubig, Barnab{\'a}s
  P{\'o}czos, and Tom~Michael Mitchell.
\newblock Competence-based curriculum learning for neural machine translation.
\newblock {\em ArXiv}, abs/1903.09848, 2019.

\bibitem{Zhang2019CurriculumLF}
Xuan Zhang, Pamela Shapiro, Manish Kumar, P.~McNamee, Marine Carpuat, and Kevin
  Duh.
\newblock Curriculum learning for domain adaptation in neural machine
  translation.
\newblock {\em ArXiv}, abs/1905.05816, 2019.

\bibitem{Penha2020CurriculumLS}
Gustavo Penha and C.~Hauff.
\newblock Curriculum learning strategies for ir.
\newblock {\em Advances in Information Retrieval}, 12035:699 -- 713, 2020.

\bibitem{xu-etal-2020-curriculum}
Benfeng Xu, Licheng Zhang, Zhendong Mao, Quan Wang, Hongtao Xie, and Yongdong
  Zhang.
\newblock Curriculum learning for natural language understanding.
\newblock In {\em Proceedings of the 58th Annual Meeting of the Association for
  Computational Linguistics}, pages 6095--6104, Online, July 2020. Association
  for Computational Linguistics.

\bibitem{Merity2016PointerSM}
Stephen Merity, Caiming Xiong, James Bradbury, and Richard Socher.
\newblock Pointer sentinel mixture models.
\newblock {\em ArXiv}, abs/1609.07843, 2016.

\bibitem{Wang2018GLUEAM}
Alex Wang, Amanpreet Singh, Julian Michael, Felix Hill, Omer Levy, and
  Samuel~R. Bowman.
\newblock Glue: A multi-task benchmark and analysis platform for natural
  language understanding.
\newblock In {\em BlackboxNLP@EMNLP}, 2018.

\bibitem{platanios-etal-2019-competence}
Emmanouil~Antonios Platanios, Otilia Stretcu, Graham Neubig, Barnabas Poczos,
  and Tom Mitchell.
\newblock Competence-based curriculum learning for neural machine translation.
\newblock In {\em Proceedings of the 2019 Conference of the North {A}merican
  Chapter of the Association for Computational Linguistics: Human Language
  Technologies, Volume 1 (Long and Short Papers)}, Minneapolis, Minnesota, June
  2019. Association for Computational Linguistics.

\bibitem{Elman1993LearningAD}
J.~Elman.
\newblock Learning and development in neural networks: the importance of
  starting small.
\newblock {\em Cognition}, 48:71--99, 1993.

\bibitem{Bengio2009CurriculumL}
Yoshua Bengio, J{\'e}r{\^o}me Louradour, Ronan Collobert, and Jason Weston.
\newblock Curriculum learning.
\newblock In {\em ICML '09}, 2009.

\bibitem{Soviany2021CurriculumLA}
Petru Soviany, Radu~Tudor Ionescu, Paolo Rota, and N.~Sebe.
\newblock Curriculum learning: A survey.
\newblock {\em ArXiv}, abs/2101.10382, 2021.

\bibitem{strubell-etal-2019-energy}
Emma Strubell, Ananya Ganesh, and Andrew McCallum.
\newblock Energy and policy considerations for deep learning in {NLP}.
\newblock In {\em Proceedings of the 57th Annual Meeting of the Association for
  Computational Linguistics}, pages 3645--3650, Florence, Italy, July 2019.
  Association for Computational Linguistics.

\bibitem{Pruksachatkun2020jiantAS}
Yada Pruksachatkun, Phil Yeres, Haokun Liu, Jason Phang, Phu~Mon Htut, Alex
  Wang, Ian Tenney, and Samuel~R. Bowman.
\newblock jiant: A software toolkit for research on general-purpose text
  understanding models.
\newblock {\em ArXiv}, abs/2003.02249, 2020.

\bibitem{Izsak2021HowTT}
Peter Izsak, Moshe Berchansky, and Omer Levy.
\newblock How to train bert with an academic budget.
\newblock {\em ArXiv}, abs/2104.07705, 2021.

\bibitem{Frankle2019TheLT}
Jonathan Frankle and Michael Carbin.
\newblock The lottery ticket hypothesis: Finding sparse, trainable neural
  networks.
\newblock {\em arXiv: Learning}, 2019.

\bibitem{Wynter2020OptimalSE}
Adrian de~Wynter and D.~Perry.
\newblock Optimal subarchitecture extraction for bert.
\newblock {\em ArXiv}, abs/2010.10499, 2020.

\bibitem{Karras2017ProgressiveGO}
Tero Karras, Timo Aila, Samuli Laine, and Jaakko Lehtinen.
\newblock Progressive growing of gans for improved quality, stability, and
  variation.
\newblock {\em ArXiv}, abs/1710.10196, 2017.

\end{thebibliography}
\section{Appendix}
\subsection{Competence Based Curricula Perplexity Results}
\begin{figure}[b]
\centering
\label{fig:wk2lp}
\includegraphics[width=10cm]{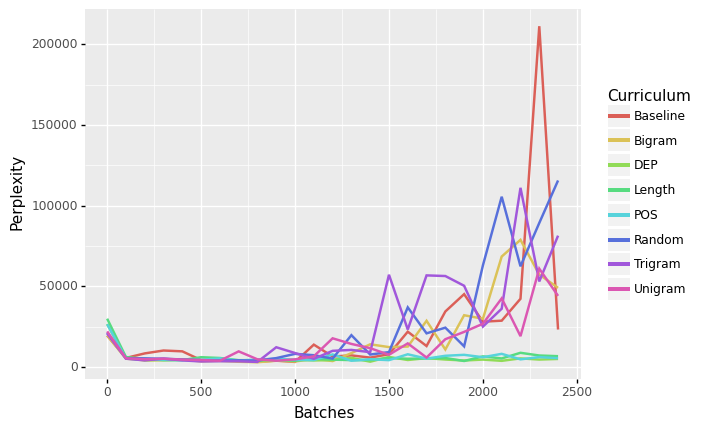}
\caption{Validation perplexity of each curriculum trained on based wikitext-2 measured every 100 batches.}
\end{figure}
\begin{figure}[t]
\centering
\label{fig:wikitext103-line}
\includegraphics[width=10cm]{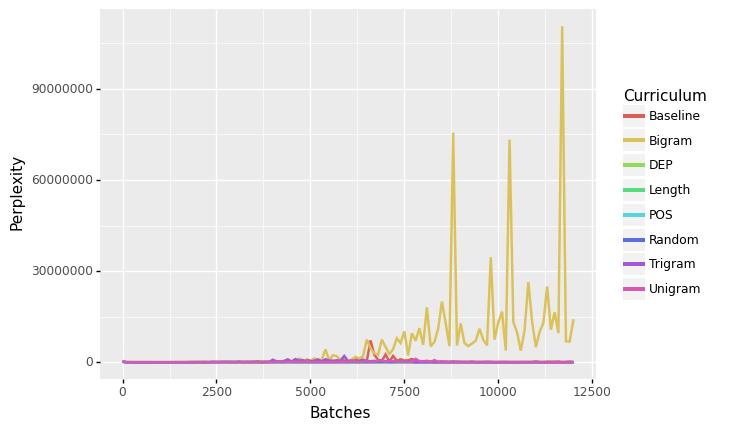}
\caption{Validation perplexity of each curriculum trained on wikitext-103 measured every 100 batches.}
\end{figure}
\begin{figure}[]
\centering
\label{fig:wikitext103-line-clean}
\includegraphics[width=10cm]{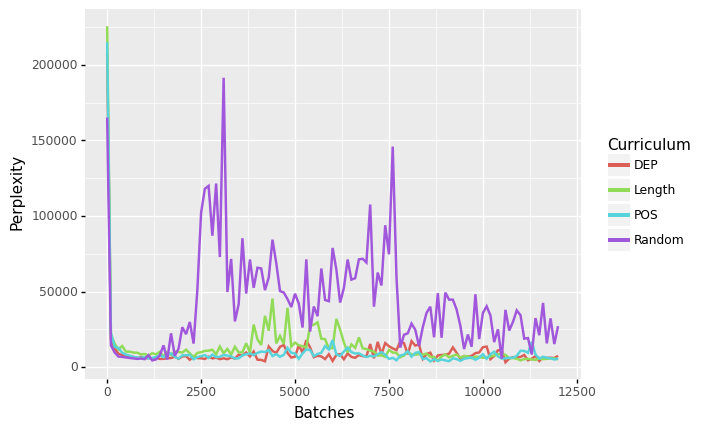}
\caption{Validation perplexity of each curriculum trained on wikitext-103 measured every 100 batches. Unigram, bigram, and baseline model performance removed improved interpretation}
\end{figure}
\begin{figure}[]
\centering
\includegraphics[width=10cm]{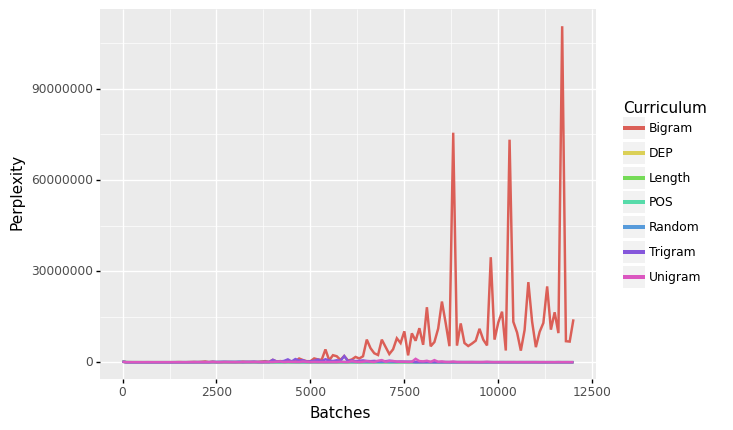}
\caption{Validation perplexity of each curriculum trained on wikitext-103 measured every 100 batches. Bigram performance is removed for ease of interpretation.}
\end{figure}
\begin{figure}[]
\centering
\includegraphics[width=10cm]{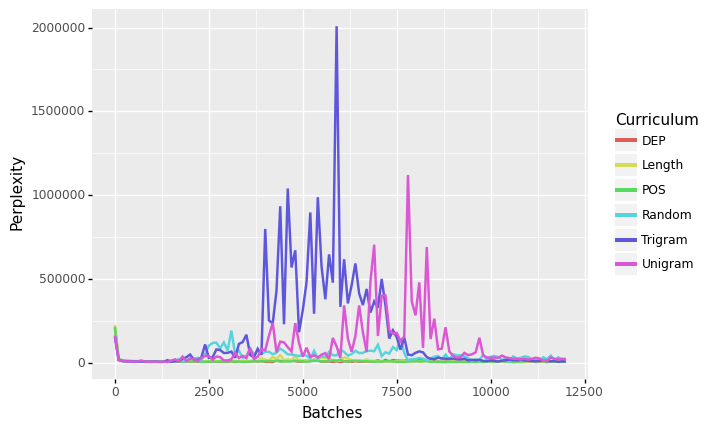}
\caption{Validation perplexity of each curriculum trained on wikitext-103 measured every 100 batches. Bigram and Baseline performance is removed for ease of interpretation.}
\end{figure}
\begin{figure}[]
\centering
\includegraphics[width=10cm]{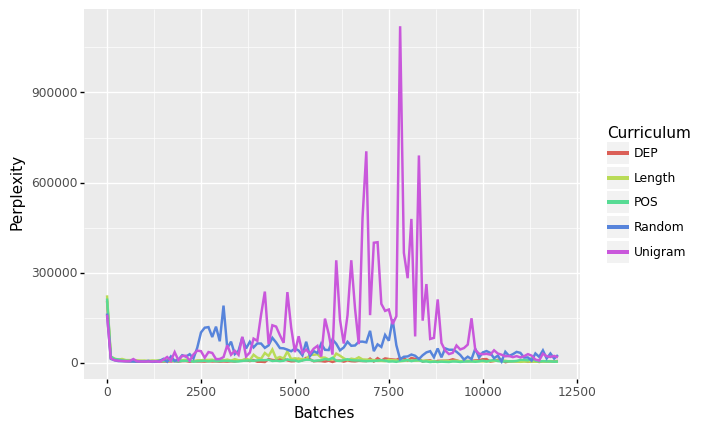}
\caption{Validation perplexity of each curriculum trained on wikitext-103 measured every 100 batches. Bigram, trigram, and Baseline performance is removed for ease of interpretation.}
\end{figure}
\begin{figure}[]
\centering
\includegraphics[width=10cm]{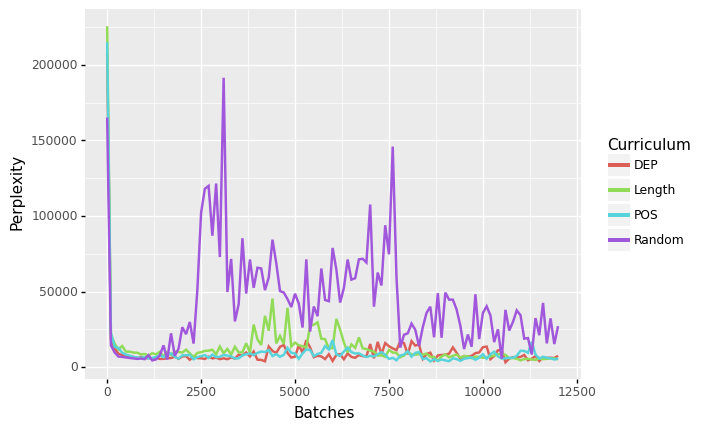}
\caption{Validation perplexity of each curriculum trained on wikitext-103 measured every 100 batches. Unigram, Bigram, Trigram and Baseline performance is removed for ease of interpretation.}
\end{figure}

\end{document}